\RequirePackage{fix-cm}
\documentclass[a4paper,10pt]{sensenova}

\usepackage{mmstyles}
\usepackage{mtforcing}

\title{MT Forcing: Towards Stable and Consistent\\
Multi-Turn Image Editing with Autoregressive Diffusion}

\author{\parbox{\textwidth}{\centering\authorfont
  \href{https://github.com/houyuanchen111}{Houyuan Chen}\textsuperscript{\ensuremath{1,2,*}},
  \href{https://scholar.google.com/citations?user=dtXIEXoAAAAJ\&hl=zh-CN}{Fanzhou Wang}\textsuperscript{\ensuremath{1,*}},
  \href{https://x-niper.github.io/}{Xiangyu Fan}\textsuperscript{\ensuremath{1,\dagger}},
  \href{https://openreview.net/profile?id=~Zesong_Qiu1}{Zesong Qiu}\textsuperscript{1},
  \href{https://scholar.google.com/citations?user=z4-wfncAAAAJ}{Yukun Wei}\textsuperscript{1},
  \href{https://songlin1998.github.io/}{Songlin Yang}\textsuperscript{2}\\[2pt]
  \href{https://scholar.google.com/citations?user=WMVmXgoAAAAJ}{Zhitao Yang}\textsuperscript{1},
  \href{https://lllyasviel.github.io/lvmin_zhang/}{Lvmin Zhang}\textsuperscript{3},
  \href{https://scholar.google.com/citations?user=muMqKAcAAAAJ\&hl=zh-CN}{Pei Yang}\textsuperscript{\ensuremath{1,2}},
  \href{https://sites.google.com/view/fromandto}{Hao Zhao}\textsuperscript{\ensuremath{4,\ddagger}},
  Dahua Lin\textsuperscript{1},
  \href{https://yanglei.me/}{Lei Yang}\textsuperscript{\ensuremath{1,\ddagger}},
  \href{https://anyirao.com/}{Anyi Rao}\textsuperscript{\ensuremath{2,\ddagger}}
}}
\contribution[*]{Equal contribution}
\contribution[\dagger]{Project lead}
\contribution[\ddagger]{Co-corresponding authors}
\affiliation[1]{SenseTime Research}
\affiliation[2]{HKUST MMLab}
\affiliation[3]{Stanford University}
\affiliation[4]{AIR, Tsinghua University}
\checkdata[Project page]{\url{https://houyuanchen111.github.io/mt-forcing.github.io/}}
\hypersetup{pdfauthor={Houyuan Chen, Fanzhou Wang, Xiangyu Fan, Zesong Qiu, Yukun Wei, Songlin Yang, Zhitao Yang, Lvmin Zhang, Pei Yang, Hao Zhao, Dahua Lin, Lei Yang, Anyi Rao}}

\abstract{
Multi-turn image editing aims to modify an image according to a sequence of instructions, enabling more effective interactive visual content creation for users. %
However, existing methods often exhibit noticeable drift after only a few editing turns, manifesting in two aspects: 1)~\textit{\textbf{unstable}} visual quality, with accumulating artifacts and degraded details; and 2)~\textit{\textbf{inconsistent}} image content, with unintended changes to preserved regions and prior edits.
To address these challenges, we propose \textbf{MT Forcing}, a novel training paradigm for multi-turn image editing with unified multimodal models~(UMMs) based on autoregressive diffusion. 
Specifically, we propose \textbf{Single-Turn} \textbf{D}istribution \textbf{M}atching \textbf{D}istillation~(\textbf{Single-Turn DMD}), which supervises student self-rollouts with a teacher that applies cumulative edits to the source image in a single turn, providing visual-quality guidance without recursive error accumulation.
Additionally, we introduce \textbf{Teacher-Forcing Guidance}, which supervises history-dependent transitions on reference trajectories during warmup and as a regularizer during distillation, encouraging cross-turn content consistency beyond what per-turn distribution matching alone can ensure.
Extensive experiments demonstrate that our approach significantly improves the multi-turn image editing capabilities of UMMs, yielding more stable visual quality and stronger cross-turn consistency.
}

\begin{document}
\maketitle
\begin{minipage}{\linewidth}
\centering
\includegraphics[width=\linewidth]{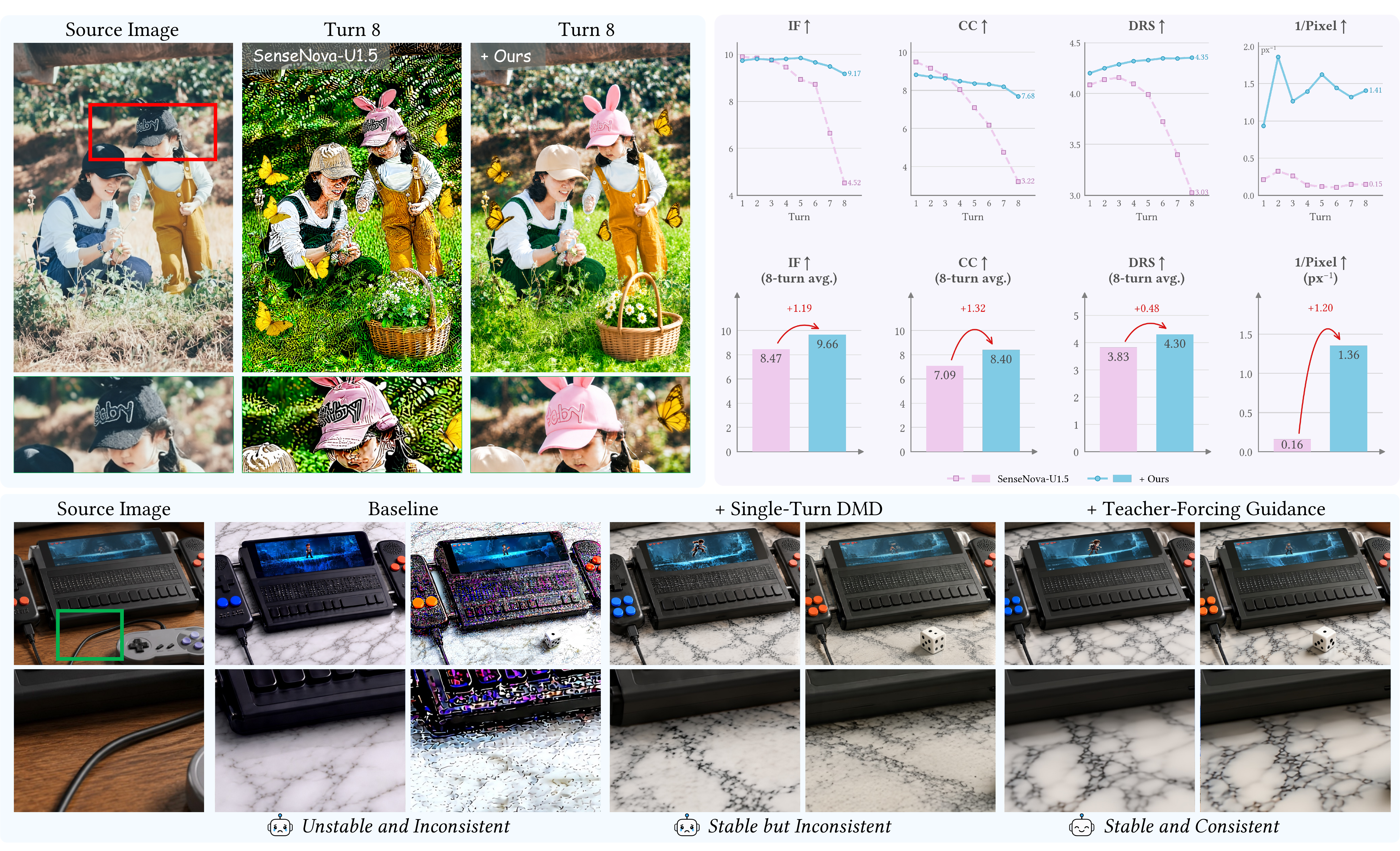}
\captionsetup{hypcap=false}
\captionof{figure}{\textbf{Overview of MT Forcing.} Top left: Our approach reduces artifacts after eight editing turns. Top right: turn-wise and eight-turn mean scores for SenseNova-U1.5 with and without MT Forcing. Bottom: qualitative comparisons illustrate the roles of Single-Turn DMD and Teacher-Forcing Guidance in improving visual quality and cross-turn consistency.}
\label{fig:teaser}
\end{minipage}\par

\clearpage

\section{Introduction}
\label{sec:intro}

Multi-turn image editing~\cite{zhou2025multi,liao2026freqedit,huang2026mt,ye2026edit,openai2026gptimage25} aims to progressively modify an image through a sequence of user instructions.
With rapid advances in unified multimodal models~(UMMs)~\cite{diao2026sensenova,sensenova2026neounify,deng2025emerging}, this task has attracted growing attention from the research community.
Compared with conventional single-turn editing~\cite{brooks2023instructpix2pix,sheynin2024emu,huang2024smartedit,li2025multiedit,mao2025visual,cui2025emu35nativemultimodalmodels,zhang2026enabling}, it enables users to continuously refine visual content as their creative intentions evolve, making it well suited to creative workflows requiring precise control over successive edits.

Since each edit builds on the results of prior turns, multi-turn image editing is inherently causal~\cite{xu2026anchoredit,qu2026vincie}, which naturally aligns with autoregressive~(AR) diffusion~\cite{guo2025long,gao2024ca2,li2024autoregressive,deng2025autoregressive} in UMMs.
Such a design enables later edits to draw on the accumulated interaction history while responding to new user instructions.
However, existing methods often exhibit drift after only a few editing turns~(see Fig.~\ref{fig:teaser} for examples), manifesting in two aspects: 1)~\textit{\textbf{unstable}} visual quality, with accumulating visual artifacts and progressive degradation of fine-grained textures and local structures; and 2)~\textit{\textbf{inconsistent}} image content, with unintended alterations to regions outside the requested edit and failures to retain prior edits that should remain unchanged. We attribute these failures to exposure bias~\cite{ning2024elucidating,schmidt2019generalization}, arising from the mismatch between training and inference conditions.
During training, the model typically conditions on reference histories, whereas inference requires recursively conditioning on its own imperfect histories.

To overcome these challenges, we propose \textbf{MT Forcing}, a training paradigm for UMMs based on AR diffusion that enables stable and consistent multi-turn image editing. First, we introduce Single-Turn \textbf{D}istribution \textbf{M}atching \textbf{D}istillation (\textbf{Single-Turn DMD}) to counteract visual-quality drift across editing turns.
Our design follows two observations: 1)~a single editing turn typically introduces limited degradation~\cite{tang2026banana100}; and 2)~editing models can execute multiple actions in one turn~\cite{team2025longcat} (see Fig.~\ref{fig:single_turn_teacher_1}).
These observations motivate using single-turn editing quality as the reference for every turn of a multi-turn student.
Specifically, we reorganize all instructions up to the current turn into a \textit{bundled prompt} that expresses their cumulative editing intent as a single edit to the source image.
Conditioned on this prompt and the source image, the teacher executes the accumulated edits in one turn, providing a quality reference without cross-turn error accumulation.
The student performs autoregressive self-rollouts during training~\cite{huang2025self,agarwal2024policy}, conditioning each edit on its own imperfect outputs.
At each turn, a DMD loss~\cite{yin2024one,yin2024improved,fan2026phased,liu2026decoupled} aligns the student's rollout output distribution with the teacher's single-turn output distribution for the same cumulative editing intent.
This trains the student to maintain visual quality under the histories encountered during inference, mitigating exposure bias and progressive visual degradation.

However, the bundled prompt specifies cumulative editing intent without fixing the visual details established in preceding student outputs~(see Fig.~\ref{fig:single_turn_teacher_2}), so per-turn distribution matching alone does not guarantee cross-turn consistency.
To improve cross-turn consistency, we introduce \textbf{Teacher-Forcing Guidance}, which trains the student with teacher forcing on consistent reference trajectories~\cite{hu2024acdit,jin2025pyramidal,zhang2026test}.
We apply this teacher-forced supervision in two stages.
1)~\textit{Teacher-Forcing Warmup}: before Single-Turn DMD training, we perform supervised fine-tuning to teach the model to produce consistent edits conditioned on coherent reference histories.
2)~\textit{Teacher-Forcing Regularization}: during Single-Turn DMD training, we retain this supervision with a lower weight as a regularizer to preserve history-dependent behavior and cross-turn consistency during distillation~\cite{you2026adaptive,wang2026pdmd,chen2026data}.

Extensive quantitative and qualitative experiments demonstrate that MT Forcing achieves more stable visual quality, greater content consistency, and better instruction following across successive editing turns.
To evaluate multi-turn image editing comprehensively, we build on MTEdit-Bench's~\cite{mtedit} assessment of instruction following, content consistency, and visual quality.
We additionally introduce \(\mathrm{Pixel}\), a metric based on image matching~\cite{he2025matchanything} that measures geometric alignment between successive editing outputs.
MT Forcing also achieves better alignment under this metric, providing additional evidence of improved content preservation across turns.

In summary, we 1)~propose Single-Turn DMD to stabilize visual quality across editing turns by aligning student rollouts with a single-turn teacher;
2)~introduce Teacher-Forcing Guidance to preserve cross-turn content consistency through teacher-forced training on consistent reference trajectories; and 3)~combine these complementary objectives into MT Forcing, which achieves more stable visual quality, greater content consistency, and better instruction following in extensive quantitative and qualitative experiments.

\section{Related Work}
\label{sec:related_work}

\paragraph{Multi-Turn Image Editing} extends instruction-based image editing~\cite{brooks2023instructpix2pix,hertz2022prompt,liu2025step1x,labs2025flux,wu2025qwenimagetechnicalreport,team2025longcat,firered2026rededit,song2026awakingspatialintelligenceunified} to an interactive setting, where users progressively refine an image through a sequence of instructions. 
A straightforward approach is to recursively apply a single-turn editor, conditioning each turn on the previous output and the current instruction~\cite{zhou2025multi,liao2026freqedit,huang2026mt,zhao2026onpolicyselfdistillationmultiturnimage}.
However, this formulation does not explicitly model the full interaction history, limiting its ability to retain earlier constraints and resolve context-dependent instructions~\cite{ye2025imgedit}.
Recent approaches build on UMMs to incorporate previous instructions and generated images into a shared context for subsequent edits~\cite{deng2025emerging,ye2026edit}.
Multi-turn Consistent Image Editing~\cite{zhou2025multi} uses flow-based inversion and dual-objective guidance to stabilize sampling, while FreqEdit~\cite{liao2026freqedit} preserves high-frequency features to reduce accumulated degradation.
VINCIE~\cite{qu2026vincie} learns in-context editing from video-derived image-text sequences, and AnchorEdit~\cite{xu2026anchoredit} introduces causal self-rollout training and memory to preserve subject identity across turns.
MT-EditFlow~\cite{huang2026mt} optimizes rewards aggregated over editing trajectories, while Edit-R2~\cite{ye2026edit} reconstructs session intent and jointly optimizes reasoning and image generation. Most recently, MT-OPSD~\cite{zhao2026onpolicyselfdistillationmultiturnimage} constructs degraded inputs through identity rollouts and distills clean-conditioned teacher predictions into the student.
However, the mismatch between training on reference histories and inference on self-generated outputs can cause errors to accumulate, making visual quality and content consistency difficult to maintain over successive turns~\cite{huang2026mt,zhao2026onpolicyselfdistillationmultiturnimage}.

\paragraph{Diffusion Distillation} aims to reduce iterative diffusion sampling to one or a few steps with minimal loss in generation quality.
Teacher-trajectory regression learns the teacher's sampling dynamics through direct noise-to-image regression~\cite{luhman2021knowledge,zheng2023fast} or progressive step compression~\cite{salimans2022progressive,meng2023distillation}, while InstaFlow~\cite{liu2023instaflow} straightens trajectories to facilitate one-step distillation. 
Consistency-based methods constrain predictions along a diffusion trajectory to share a common endpoint~\cite{song2023consistency,luo2023latent,luo2023lcm}, with extensions to interval mappings~\cite{kim2024consistency,zheng2024trajectory,heek2024multistep,wang2024phased} and transitive or bootstrapped supervision~\cite{berthelot2023tract,gu2023boot}.
Continuous-time formulations improve scalability~\cite{lu2025simplifying}, while rCM~\cite{zheng2026large} combines consistency distillation with score regularization.
Distribution-level methods supervise generated samples through adversarial objectives~\cite{sauer2024adversarial,sauer2024fast,lin2024sdxl,xu2024ufogen,kang2024distilling} or score-based matching~\cite{luo2023diff,nguyen2024swiftbrush,zhou2024score}.
DMD~\cite{yin2024one} matches distributions using the difference between target and student-distribution scores, while DMD2~\cite{yin2024improved} stabilizes online score estimation and incorporates adversarial supervision.
Subsequent work improves distillation quality and scalability~\cite{dao2024swiftbrush,zhou2024adversarial,lu2025adversarial,ge2025senseflow}, refines multi-step matching and guidance~\cite{fan2026phased,wu2026diversity,liu2026decoupled}, and explores alternative divergence objectives~\cite{xu2025one}.
Auxiliary student-score estimation can also be replaced by generator-derived surrogates~\cite{kim2026distribution} or discriminative density-ratio estimation~\cite{yu2026dmad}.
We use DMD2 to transfer single-turn editing quality to outputs under imperfect multi-turn histories.

\paragraph{Autoregressive Diffusion} generates visual content in a causal autoregressive manner.
For images, UMMs~\cite{deng2025emerging,sensenova2026neounify,diao2026sensenova,diao2026sensenovau15nativeunifiedvisual,xie2025show,xie2026show,ge2024seed,fu2026lanceunifiedmultimodalmodeling} extend sequence modeling to visual understanding, generation, and editing, often combining autoregressive context modeling with diffusion-based image synthesis.
For videos, a common strategy distills pretrained bidirectional diffusion models into few-step autoregressive generators, reducing latency and enabling streaming beyond the training clip length~\cite{yin2025slow,huang2025self}.
CausVid~\cite{yin2025slow} uses asymmetric distribution matching between a bidirectional teacher and a causal student.
Diffusion Forcing~\cite{chen2024diffusion} trains with independently noised sequence elements, whereas Self Forcing~\cite{huang2025self} trains on autoregressive self-rollouts to align training with inference.
Other studies improve causal student initialization~\cite{zhu2026causal,zheng2026causal} and scalable few-step or one-step distillation~\cite{zhao2026causal,feng2026one}.
Long-horizon stability is further strengthened through streaming training and error control~\cite{cui2026self,liu2026rolling,yang2025longlive,li2025stable,zhu2026stream,cao2026recency}, long-range context modeling~\cite{chen2026context,xue2026ring,xu2026recap}, and KV-cache management~\cite{yi2025deep,zhao2026relax,luo2026future,shi2026surprise}.
Further work improves distillation quality and diversity~\cite{lu2026reward,zhao2026mask} and reduces inference costs through sparse or multi-scale attention~\cite{lv2026light,xu2026sparse,li2026ms}.
However, directly transferring these methods to multi-turn image editing is challenging, as bidirectional teachers capable of jointly generating complete editing trajectories are unavailable.

\section{Method}
\label{sec:method}

\begin{figure}[tbp]
    \centering
    \includegraphics[width=\textwidth]{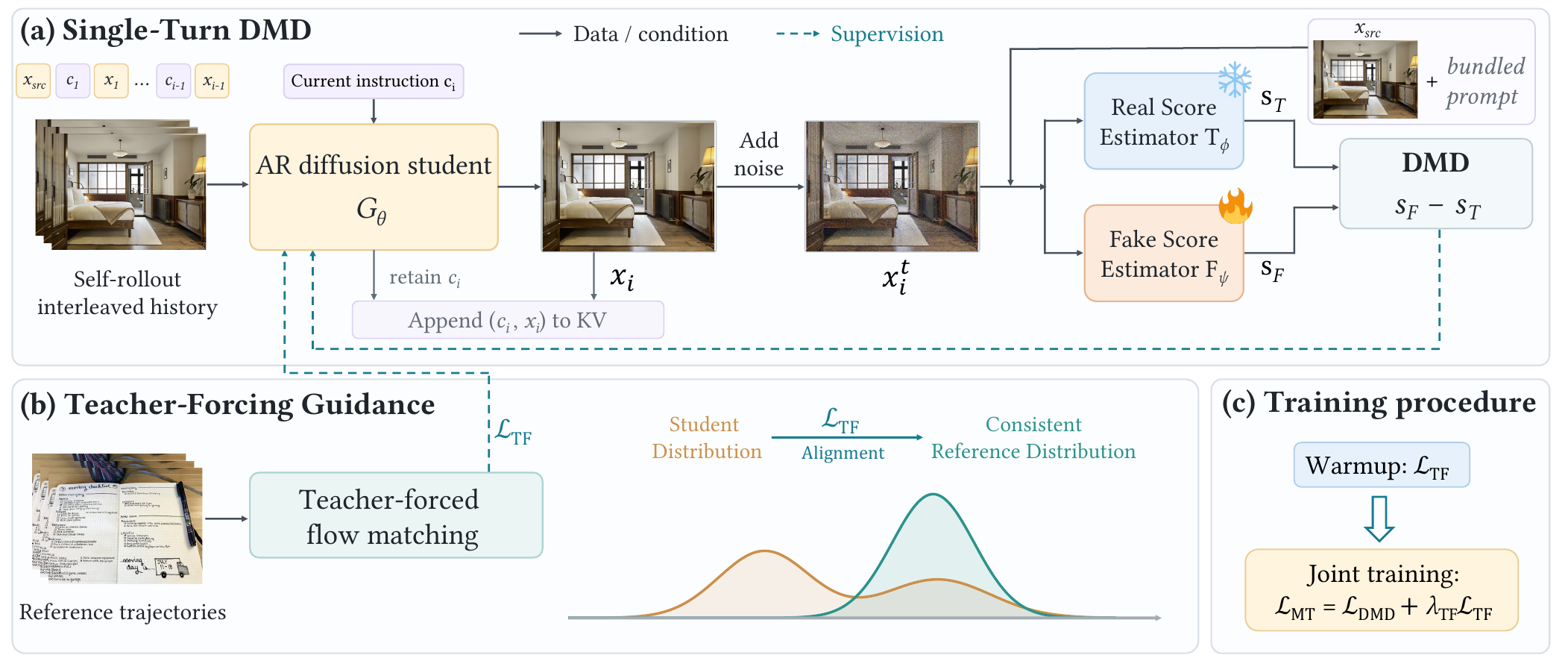}
    \caption{\textbf{Overview of MT Forcing.} (a)~Single-Turn DMD matches student rollout marginals to single-turn teacher distributions. (b)~Teacher-Forcing Guidance supervises history-dependent transitions on reference trajectories via flow matching to encourage cross-turn consistency. (c)~Overall training: teacher-forcing warmup followed by DMD with teacher-forcing regularization.}
    \label{fig:mt_forcing_framework}
\end{figure}

As shown in Fig.~\ref{fig:mt_forcing_framework}, MT Forcing combines distribution matching on self-generated histories with supervision on reference trajectories.
Sec.~\ref{sec:method_formulation} defines the multi-turn editing problem and its training--inference mismatch.
Sec.~\ref{sec:single_turn_dmd} introduces Single-Turn DMD to supervise student rollouts with a single-turn teacher.
Sec.~\ref{sec:teacher_forcing_guidance} presents Teacher-Forcing Guidance for cross-turn consistency.
Algorithm~\ref{alg:mt-forcing} summarizes the overall training procedure.

\subsection{Problem Formulation}
\label{sec:method_formulation}
A multi-turn editing session consists of a source image $x_{\mathrm{src}}$ and $N$ instructions $c_{1:N}$.
At turn $i$, the model $G_\theta$ generates $x_i$ from the current instruction $c_i$ and the self-rollout interleaved history $H_{<i}=(x_{\mathrm{src}},c_1,x_1,\ldots,c_{i-1},x_{i-1})$.
\begin{equation}
    p_\theta(x_{1:N}\mid x_{\mathrm{src}},c_{1:N})
    =\prod_{i=1}^{N}p_\theta(x_i\mid H_{<i},c_i).
    \label{eq:mt_autoregressive}
\end{equation}
The causal model caches key--value (KV) vectors of previous instructions and images for subsequent edits~\cite{teng2025magi,lin2026autoregressive}.

In multi-turn image editing, UMMs encounter different histories during teacher-forcing training and rollout inference.
A reference trajectory $\tau=(x_{\mathrm{src}},\{(c_i,x_i^\star)\}_{i=1}^{N})\sim\mathcal D_{\mathrm{ref}}$ supplies $H_{<i}^\star=(x_{\mathrm{src}},c_1,x_1^\star,\ldots,c_{i-1},x_{i-1}^\star)$ during teacher forcing, whereas during inference the model uses its own preceding outputs:
\begin{equation}
    \begin{aligned}
        \text{Teacher-forcing training:}\quad
        &p_\theta(x_i\mid H_{<i}^\star,c_i),\\
        \text{Self-rollout inference:}\quad
        &p_\theta(x_i\mid H_{<i},c_i).
    \end{aligned}
    \label{eq:mt_train_test_gap}
\end{equation}
Therefore, errors in generated histories can accumulate across turns, causing editing results to drift progressively.

\subsection{Single-Turn Distribution Matching Distillation}
\label{sec:single_turn_dmd}

Recent video-generation methods~\cite{huang2025self,cui2026self} mitigate rollout drift by distilling from pretrained bidirectional teachers that jointly model multiple frames.
Multi-turn editing lacks a corresponding bidirectional UMM that jointly generates instruction-conditioned editing trajectories.
We therefore propose Single-Turn DMD, in which a single-turn editing teacher supervises each generated turn; we next specify its conditioning and training signal.

At turn $i$, we reorganize all editing instructions up to that turn, $c_{1:i}$, into a single-turn editing prompt, termed the \textit{bundled prompt} $\bar c_i=\operatorname{Reorganize}(c_{1:i})$.
This prompt expresses the cumulative editing intent as one edit to be applied directly to the source image.
For example, ``add a vase'' followed by ``make it blue'' can be reorganized as ``add a blue vase.''
We use $\mathcal C_i^{\mathrm S}=(H_{<i},c_i)$ and $\mathcal C_i^{\mathrm T}=(x_{\mathrm{src}},\bar c_i)$ as the student and teacher conditions, respectively.
The student recursively samples $x_i\sim p_\theta(\cdot\mid\mathcal C_i^{\mathrm S})$ and appends $(c_i,x_i)$ to its history.
Since $\mathcal C_i^{\mathrm S}$ contains the source image and all instructions up to turn $i$, prompt reorganization maps it to $\mathcal C_i^{\mathrm T}$, but this mapping is not invertible:
\begin{equation}
    \mathcal C_i^{\mathrm S}\Rightarrow\mathcal C_i^{\mathrm T},
    \qquad
    \mathcal C_i^{\mathrm T}\nRightarrow\mathcal C_i^{\mathrm S}.
    \label{eq:mt_condition_mapping}
\end{equation}
Thus, the student receives \emph{stronger conditioning} and the frozen teacher $T_\phi$ receives \emph{weaker conditioning}.

\begin{figure}[tbp]
    \centering
    \begin{minipage}[t]{0.49\linewidth}
        \vspace{0pt}
        \centering
        \includegraphics[width=\linewidth]{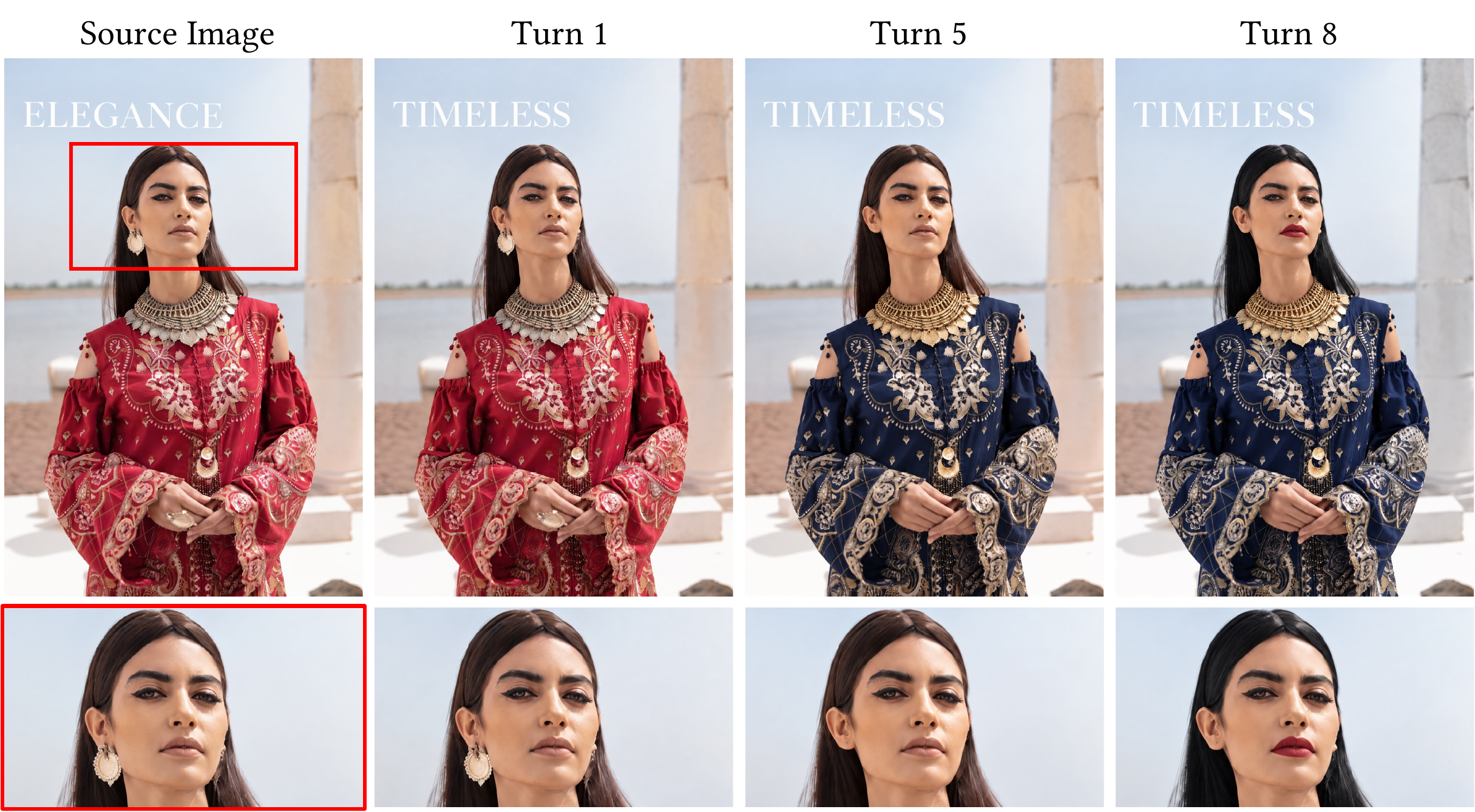}
        \caption{\textbf{Visual quality across turns with a single-turn teacher.} Each bundled prompt is applied independently to the source image. Enlarged crops show stable facial details.}
        \label{fig:single_turn_teacher_1}
    \end{minipage}\hfill
    \begin{minipage}[t]{0.49\linewidth}
        \vspace{0pt}
        \centering
        \includegraphics[width=\linewidth]{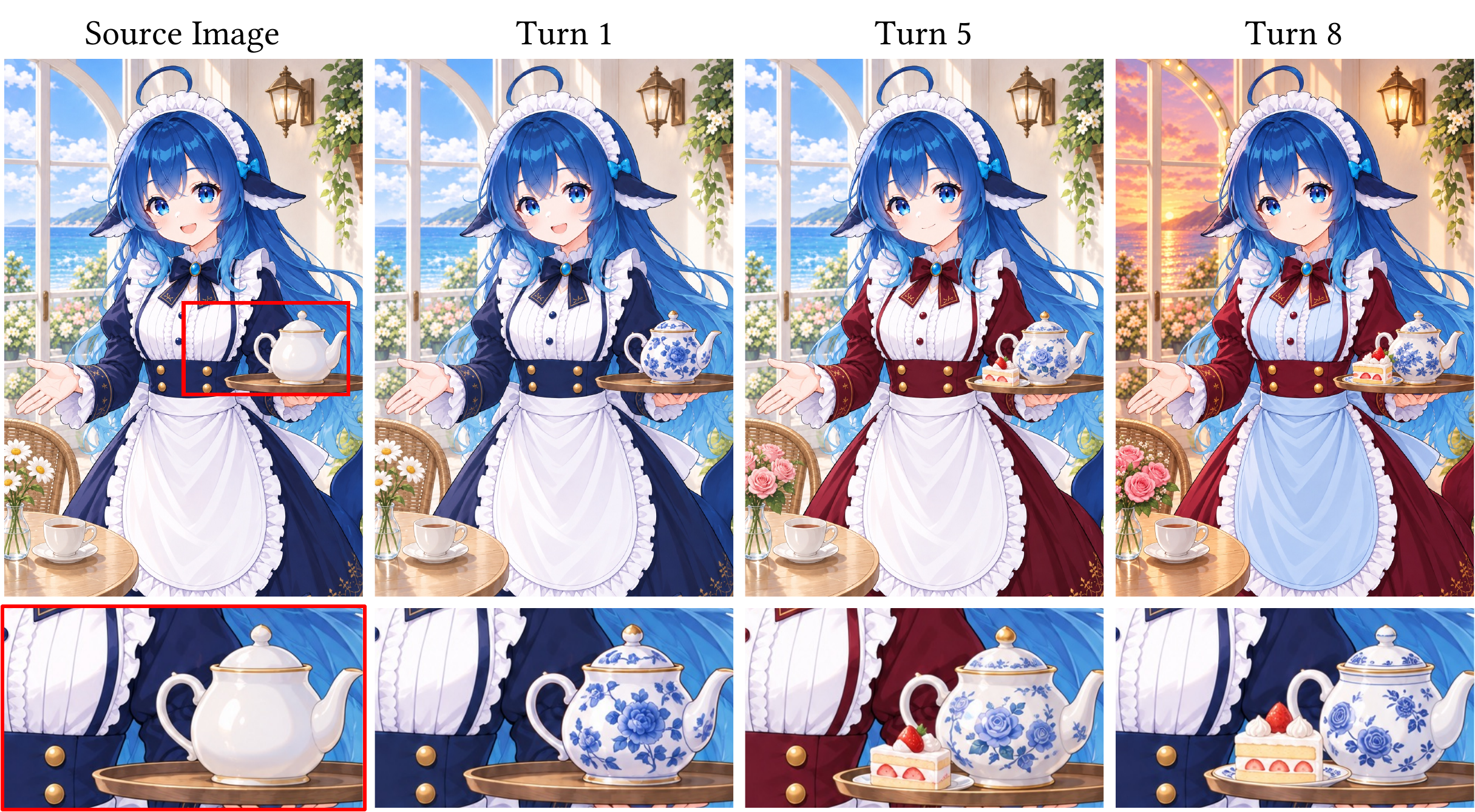}
        \caption{\textbf{Content inconsistency of the single-turn teacher.} The teapot's floral pattern changes across independent edits with bundled prompts, as highlighted in the enlarged crops.}
        \label{fig:single_turn_teacher_2}
    \end{minipage}
\end{figure}

Fig.~\ref{fig:single_turn_teacher_1} shows clear teacher outputs at Turns~1, 5, and~8 for bundled prompts formed from accumulated editing instructions.
By realizing cumulative editing intent in a single turn while maintaining visual quality across editing stages, the teacher can therefore provide effective per-turn distributional supervision for the student.

\paragraph{\textit{How can a teacher with weaker conditioning train the student?}}
The connection is provided by the distribution induced by the student's causal rollout.
The student generates each image from $p_\theta(\cdot\mid\mathcal C_i^{\mathrm S})$, using the accumulated history and the current instruction.
For fixed $\mathcal C_i^{\mathrm T}$, let $p_{\theta,i}(\cdot\mid\mathcal C_i^{\mathrm T})$ denote the marginal distribution of $x_i$ under the student's rollout, averaged over instruction sequences consistent with $\mathcal C_i^{\mathrm T}$.
\begin{equation}
    p_{\theta,i}(x_i\mid\mathcal C_i^{\mathrm T})
    =\mathbb E_{c_{1:i}\mid\mathcal C_i^{\mathrm T}}\!\left[
    \int p_\theta(x_i\mid\mathcal C_i^{\mathrm S})
    \prod_{j<i}p_\theta(x_j\mid\mathcal C_j^{\mathrm S})
    \,d x_{1:i-1}\right].
    \label{eq:mt_rollout_marginal}
\end{equation}

Following DMD~\cite{luo2023diff,yin2024one,yin2024improved}, we apply the same forward process $x_i^t=\alpha_t x_i+\sigma_t\epsilon$, $\epsilon\sim\mathcal N(0,I)$, to both distributions.
Image variables may denote pixels or a shared latent representation.
Writing the resulting distributions as $p_{\theta,i,t}$ and $p_{\phi,i,t}$, we define the per-turn loss for a fixed $\mathcal C_i^{\mathrm T}$ as
\begin{equation}
    \mathcal L_{\mathrm{DMD}}^{(i)}
    =\mathbb E_t\!\left[
        D_{\mathrm{KL}}\!\left(
            p_{\theta,i,t}(\cdot\mid\mathcal C_i^{\mathrm T})
            \,\middle\|\,
            p_{\phi,i,t}(\cdot\mid\mathcal C_i^{\mathrm T})
        \right)
    \right],
    \label{eq:mt_dmd_per_turn}
\end{equation}
where $t$ follows a fixed distribution over nondegenerate noise levels.
The KL compares the two noisy-image distributions under the same source image and bundled prompt.
Through Eq.~\ref{eq:mt_rollout_marginal}, optimizing this loss updates the student that actually generates under $\mathcal C_i^{\mathrm S}$, using a distribution-matching target expressed under $\mathcal C_i^{\mathrm T}$.
The full objective averages these losses over turns and task conditions induced by sampled instruction sequences.

In practice, we optimize Eq.~\ref{eq:mt_dmd_per_turn} through a score-based gradient estimate~\cite{yin2024one,yin2024improved}:
\begin{equation}
    \nabla_\theta\mathcal L_{\mathrm{DMD}}^{(i)}
    \approx\mathbb E\!\left[
        \left(\frac{\partial x_i^t}{\partial\theta}\right)^{\!\top}
        \bigl(\hat{\mathbf s}_i^{\mathrm F}-\hat{\mathbf s}_i^{\mathrm T}\bigr)
    \right].
    \label{eq:mt_dmd_gradient}
\end{equation}
Here $\hat{\mathbf s}_i^{\mathrm T}$ is provided by the frozen teacher $T_\phi$, while $\hat{\mathbf s}_i^{\mathrm F}$ is estimated by the online fake model $F_\psi$, trained to denoise detached rollout outputs; both are conditioned on $\mathcal C_i^{\mathrm T}$.
The student is updated by backpropagating this score difference through its generated image, treating the score estimates as fixed.

For efficiency, we truncate gradients across turns by detaching historical KV embeddings and retain gradients only through the sampled denoising step $s_i$ of the current turn~\cite{huang2025self}.
Thus, the partial derivative in Eq.~\ref{eq:mt_dmd_gradient} holds the history and incoming denoising state fixed.
We denote the single-sample update direction inside the expectation by $\widehat{\mathbf g}_i$ and average these directions over turns for one student update after each rollout (Algorithm~\ref{alg:mt-forcing}).

\subsection{Teacher-Forcing Guidance}
\label{sec:teacher_forcing_guidance}

\paragraph{\textit{Why is Single-Turn DMD not enough?}}
Single-Turn DMD trains the history-conditioned student through its rollout marginal, but matching this marginal does not uniquely determine the student's response to each history.
The objective constrains the output distribution given $\mathcal C_i^{\mathrm T}$, leaving different dependencies on the additional visual history in $\mathcal C_i^{\mathrm S}$ indistinguishable.
To see this, consider a student that ignores preceding generated images and independently samples from the teacher at each turn:
\begin{equation}
    p_\theta(x_i\mid H_{<i},c_i)
    =p_{\phi,i}(x_i\mid\mathcal C_i^{\mathrm T}).
    \label{eq:mt_independent_solution}
\end{equation}
Under Eq.~\ref{eq:mt_independent_solution}, Eq.~\ref{eq:mt_rollout_marginal} gives $p_{\theta,i}=p_{\phi,i}$ and zero ideal DMD loss at every turn.
Nevertheless, independent sampling can change content across turns, even when every per-turn marginal distribution matches the teacher's.
Thus, zero DMD loss does not guarantee consistent trajectories.
The changing teapot pattern in Fig.~\ref{fig:single_turn_teacher_2} illustrates this ambiguity.

We introduce Teacher-Forcing Guidance to promote cross-turn consistency using reference trajectories $\tau\sim\mathcal D_{\mathrm{ref}}$.
Given the reference history $H_{<i}^\star$ and current instruction $c_i$, we supervise the target $x_i^\star$ through flow matching~\cite{lipman2022flow}:
\begin{equation}
    \mathcal L_{\mathrm{TF}}
    =\frac{1}{N}\sum_{i=1}^{N}\mathbb E_{\tau,t_i,\epsilon_i}
    \Bigl[\bigl\|v_\theta(x_i^{\star,t_i},t_i,H_{<i}^\star,c_i)
    -v_i^{\star,t_i}\bigr\|_2^2\Bigr],
    \label{eq:mt_tf_flow_matching}
\end{equation}
where $v_\theta$ is the student's velocity prediction, $x_i^{\star,t_i}=\alpha_{t_i}x_i^\star+\sigma_{t_i}\epsilon_i$, and $v_i^{\star,t_i}=\dot\alpha_{t_i}x_i^\star+\dot\sigma_{t_i}\epsilon_i$, with $\epsilon_i\sim\mathcal N(0,I)$ and dots denoting time derivatives.
Pairing each target with its reference history encourages the model to preserve content across turns while following the current instruction.
This complements Single-Turn DMD's marginal distribution matching by supervising history--output dependencies.
We use $\mathcal L_{\mathrm{TF}}$ in two stages: teacher-forcing warmup before distillation and teacher-forcing regularization during distillation.

\begin{wrapfloat}{algorithm}[29]{R}{0.55\textwidth}
    \setstretch{1}
\setlength{\parskip}{0pt}
\captionsetup{font=footnotesize,labelfont={bf,color=mtink},labelsep=space,skip=4pt}
\caption{MT Forcing training procedure}
\label{alg:mt-forcing}
\fontsize{8}{9.3}\selectfont
\algrenewcommand\algorithmicindent{0.8em}
\begin{algorithmic}[1]
    \Require Denoising timesteps $\{t_1,\ldots,t_T\}$
    \Require Reference editing trajectories $\mathcal D_{\mathrm{ref}}$
    \Require Pretrained editing model $G_{\theta_0}$
    \Require Number of editing turns $N$
    \State Warm up $G_{\theta_0}$ on $\mathcal D_{\mathrm{ref}}$ using $\mathcal L_{\mathrm{TF}}$ (Eq.~\ref{eq:mt_tf_flow_matching}) to obtain $\theta_{\mathrm w}$
    \State Initialize $G_\theta$, $T_\phi$, and $F_\psi$ with weights $\theta_{\mathrm w}$
    \State Freeze teacher $T_\phi$
    \Loop
        \State Sample $\tau=(x_{\mathrm{src}},\{(c_i,x_i^\star)\}_{i=1}^{N})\sim\mathcal D_{\mathrm{ref}}$
        \State Initialize model output $\mathbf X_\theta\gets[\,]$
        \State Disable gradient computation
        \State Initialize KV cache $\mathbf{KV}\gets[G_\theta^{\mathrm{KV}}(x_{\mathrm{src}};0)]$
        \For{$i=1,\ldots,N$}
            \State Reorganize $c_{1:i}$ into a single-turn prompt $\bar c_i$ and set $\mathcal C_i^{\mathrm T}=(x_{\mathrm{src}},\bar c_i)$
            \State Sample $s_i\sim\operatorname{Uniform}(1,2,\ldots,T)$
            \State Initialize $x_i^{t_T}\sim\mathcal N(0,I)$
            \For{$j=T,\ldots,s_i$}
                \If{$j=s_i$}
                    \State Enable gradient computation
                    \State Set $\hat x_i^0\gets G_\theta(x_i^{t_j};t_j,\mathbf{KV},c_i)$
                    \State $\mathbf X_\theta\texttt{.append}(\hat x_i^0)$
                    \State Disable gradient computation
                    \State Cache $\mathbf{kv}_i\gets G_\theta^{\mathrm{KV}}((c_i,\hat x_i^0);0,\mathbf{KV})$
                    \State $\mathbf{KV}\texttt{.append}(\mathbf{kv}_i)$
                \Else
                    \State Disable gradient computation
                    \State Set $\hat x_i^0\gets G_\theta(x_i^{t_j};t_j,\mathbf{KV},c_i)$
                    \State Sample $\epsilon\sim\mathcal N(0,I)$
                    \State Set $x_i^{t_{j-1}}\gets\Psi(\hat x_i^0,\epsilon,t_{j-1})$
                \EndIf
            \EndFor
            \State Compute truncated $\widehat{\mathbf g}_i$ using $T_\phi,F_\psi$ under $\mathcal C_i^{\mathrm T}$ (Eq.~\ref{eq:mt_dmd_gradient})
        \EndFor
        \State Update $\theta$ using $N^{-1}\sum_i\widehat{\mathbf g}_i+\lambda_{\mathrm{TF}}\nabla_\theta\mathcal L_{\mathrm{TF}}$
        \State Update $\psi$ by denoising detached $\mathbf X_\theta$ with the corresponding $\mathcal C_i^{\mathrm T}$
    \EndLoop
\end{algorithmic}

\end{wrapfloat}
\paragraph{Teacher-Forcing Warmup.}
Before distillation, we optimize $\mathcal L_{\mathrm{TF}}$ to adapt the pretrained editor to reference multi-turn contexts.
The checkpoint $\theta_{\mathrm w}$ initializes the student $G_\theta$, frozen teacher $T_\phi$, and online fake model $F_\psi$.
Warmup establishes history-dependent behavior, but an initialization does not constrain subsequent distribution-matching updates to retain it.

\paragraph{Teacher-Forcing Regularization.}
We retain reference supervision during distillation to penalize changes to reference-conditioned transitions that the marginal objective cannot distinguish.
The combined objective is
\begin{equation}
    \mathcal L_{\mathrm{MT}}
    =\mathcal L_{\mathrm{DMD}}
    +\lambda_{\mathrm{TF}}\mathcal L_{\mathrm{TF}},
    \label{eq:mt_joint_objective}
\end{equation}
The update combines the averaged truncated DMD directions with $\lambda_{\mathrm{TF}}\nabla_\theta\mathcal L_{\mathrm{TF}}$.
DMD supervises the output distribution under weaker conditioning on rollouts, while Teacher-Forcing Regularization anchors history-dependent behavior on reference trajectories.

\section{Experiments}
\label{sec:experiments}
\suppressfloats[t]
In this section, we present experimental results for MT Forcing on multi-turn image editing.
We first describe the baselines, benchmark, and evaluation metrics in Sec.~\ref{sec:experimental_setup}.
We then present quantitative and qualitative comparisons in Sec.~\ref{sec:main_results}.
Finally, in Sec.~\ref{sec:ablation_studies}, we conduct ablation studies to assess the contribution of each component in MT Forcing.

\subsection{Experimental Setup}
\label{sec:experimental_setup}

For baselines, we evaluate MT Forcing on two UMMs: SenseNova-U1~\cite{diao2026sensenova} and SenseNova-U1.5~\cite{diao2026sensenovau15nativeunifiedvisual}.
We compare each editor's performance with and without MT Forcing.
Ablation studies use SenseNova-U1.5~\cite{diao2026sensenovau15nativeunifiedvisual} as the baseline.

For benchmarking, we evaluate on the step-wise sub-bench of MTEdit-Bench~\cite{mtedit}, which contains 100 sessions, each comprising a source image and eight sequential editing instructions.

For metrics, we follow MTEdit-Bench~\cite{mtedit} and report Instruction Following~(IF), Content Consistency~(CC), and Drift Robustness Score~(\(\mathrm{DRS}\)).
IF measures fulfillment of the current instruction, while CC evaluates preservation of the intended cumulative content under all instructions up to the current turn.
Both are scored from 0 to 10 by Qwen3.5-Plus~\cite{qwen35blog}.
\(\mathrm{DRS}\) measures robustness against visual-quality drift across turns, computed as the mean of the VisualQuality-R1~\cite{wu2026visualquality} and RALI~\cite{zhao2025reasoning} scores, and ranges from 0 to 5.
Higher values indicate better performance for all three metrics.
Moreover, we introduce \(\mathrm{Pixel}\) to quantify geometric alignment between successive editing outputs.
Using MatchAnything-RoMa~\cite{he2025matchanything}, we compute the mean matched-point displacement in pixels as $\mathrm{Pixel}_i=|\mathcal M_i|^{-1}\sum_{(u,v)\in\mathcal M_i}\|u-v\|_2$, where $\mathcal M_i$ contains valid matches between $x_{i-1}$ and $x_i$, with $x_0=x_{\mathrm{src}}$.
Lower values indicate less geometric change in matched regions.

\subsection{Main Results}
\label{sec:main_results}

\begin{table}[t]
\captionsetup{position=top,skip=6pt}
\caption{\textbf{Quantitative results on the step-wise sub-bench of MTEdit-Bench.} Results compare each base editor with MT Forcing across eight turns. Higher IF, CC, and \(\mathrm{DRS}\), and lower \(\mathrm{Pixel}\), indicate better performance.}
\label{tab:mt_step_wise_results}
\centering
\newcommand{\mtwithdelta}[3]{#1\raisebox{-1.4pt}{{\fontsize{4.5}{4.5}\selectfont\textcolor{#2}{#3}}}}
\fontsize{7.5}{8}\selectfont
\setlength{\tabcolsep}{0.25pt}
\renewcommand{\arraystretch}{1.0}
\setlength{\aboverulesep}{1.2pt}
\setlength{\belowrulesep}{1.2pt}
\begin{tabular*}{\linewidth}{@{\extracolsep{\fill}}l@{\hspace{1pt}}*{3}{cccc@{\hspace{1pt}}}cccc@{}}
\toprule
\textbf{Method} & \multicolumn{4}{c}{\textbf{Turn 1}} & \multicolumn{4}{c}{\textbf{Turn 2}} & \multicolumn{4}{c}{\textbf{Turn 3}} & \multicolumn{4}{c}{\textbf{Turn 4}} \\
\cmidrule(lr){2-5}\cmidrule(lr){6-9}\cmidrule(lr){10-13}\cmidrule(lr){14-17}
& IF$\uparrow$ & CC$\uparrow$ & \(\mathrm{DRS}\uparrow\) & \(\mathrm{Pixel}\downarrow\) & IF$\uparrow$ & CC$\uparrow$ & \(\mathrm{DRS}\uparrow\) & \(\mathrm{Pixel}\downarrow\) & IF$\uparrow$ & CC$\uparrow$ & \(\mathrm{DRS}\uparrow\) & \(\mathrm{Pixel}\downarrow\) & IF$\uparrow$ & CC$\uparrow$ & \(\mathrm{DRS}\uparrow\) & \(\mathrm{Pixel}\downarrow\) \\
\midrule
SenseNova-U1~\cite{diao2026sensenova} & 9.29 & \textbf{9.06} & 4.07 & 6.63 & 8.95 & 8.51 & 4.14 & 13.19 & 8.53 & 7.61 & 4.12 & 12.28 & 7.37 & 5.95 & 3.90 & 3.37 \\
\rowcolor{mtoursbg}\hspace{6pt}\textbf{+ MT Forcing} & \mtwithdelta{\textbf{9.71}}{green!45!black}{+.42} & \mtwithdelta{8.85}{red!75!black}{-.21} & \mtwithdelta{\textbf{4.25}}{green!45!black}{+.18} & \mtwithdelta{\textbf{1.23}}{green!45!black}{-5.40} & \mtwithdelta{\textbf{9.78}}{green!45!black}{+.83} & \mtwithdelta{\textbf{8.55}}{green!45!black}{+.04} & \mtwithdelta{\textbf{4.23}}{green!45!black}{+.09} & \mtwithdelta{\textbf{0.39}}{green!45!black}{-12.80} & \mtwithdelta{\textbf{9.65}}{green!45!black}{+1.12} & \mtwithdelta{\textbf{8.57}}{green!45!black}{+.96} & \mtwithdelta{\textbf{4.23}}{green!45!black}{+.11} & \mtwithdelta{\textbf{0.38}}{green!45!black}{-11.90} & \mtwithdelta{\textbf{9.56}}{green!45!black}{+2.19} & \mtwithdelta{\textbf{8.47}}{green!45!black}{+2.52} & \mtwithdelta{\textbf{4.25}}{green!45!black}{+.35} & \mtwithdelta{\textbf{0.35}}{green!45!black}{-3.02} \\
\midrule
SenseNova-U1.5~\cite{diao2026sensenovau15nativeunifiedvisual} & \textbf{9.90} & \textbf{9.49} & 4.09 & 4.74 & \textbf{9.85} & \textbf{9.16} & 4.14 & 3.10 & 9.76 & \textbf{8.76} & 4.16 & 3.86 & 9.46 & 8.04 & 4.09 & 7.30 \\
\rowcolor{mtoursbg}\hspace{6pt}\textbf{+ MT Forcing} & \mtwithdelta{9.74}{red!75!black}{-.16} & \mtwithdelta{8.82}{red!75!black}{-.67} & \mtwithdelta{\textbf{4.20}}{green!45!black}{+.11} & \mtwithdelta{\textbf{1.07}}{green!45!black}{-3.67} & \mtwithdelta{9.81}{red!75!black}{-.04} & \mtwithdelta{8.71}{red!75!black}{-.45} & \mtwithdelta{\textbf{4.25}}{green!45!black}{+.11} & \mtwithdelta{\textbf{0.54}}{green!45!black}{-2.56} & \mtwithdelta{\textbf{9.77}}{green!45!black}{+.01} & \mtwithdelta{8.64}{red!75!black}{-.12} & \mtwithdelta{\textbf{4.29}}{green!45!black}{+.13} & \mtwithdelta{\textbf{0.79}}{green!45!black}{-3.07} & \mtwithdelta{\textbf{9.81}}{green!45!black}{+.35} & \mtwithdelta{\textbf{8.49}}{green!45!black}{+.45} & \mtwithdelta{\textbf{4.32}}{green!45!black}{+.23} & \mtwithdelta{\textbf{0.72}}{green!45!black}{-6.58} \\
\midrule
\textbf{Method} & \multicolumn{4}{c}{\textbf{Turn 5}} & \multicolumn{4}{c}{\textbf{Turn 6}} & \multicolumn{4}{c}{\textbf{Turn 7}} & \multicolumn{4}{c}{\textbf{Turn 8}} \\
\cmidrule(lr){2-5}\cmidrule(lr){6-9}\cmidrule(lr){10-13}\cmidrule(lr){14-17}
& IF$\uparrow$ & CC$\uparrow$ & \(\mathrm{DRS}\uparrow\) & \(\mathrm{Pixel}\downarrow\) & IF$\uparrow$ & CC$\uparrow$ & \(\mathrm{DRS}\uparrow\) & \(\mathrm{Pixel}\downarrow\) & IF$\uparrow$ & CC$\uparrow$ & \(\mathrm{DRS}\uparrow\) & \(\mathrm{Pixel}\downarrow\) & IF$\uparrow$ & CC$\uparrow$ & \(\mathrm{DRS}\uparrow\) & \(\mathrm{Pixel}\downarrow\) \\
\midrule
SenseNova-U1~\cite{diao2026sensenova} & 6.21 & 4.41 & 3.55 & 11.06 & 4.50 & 3.06 & 3.14 & 2.42 & 2.96 & 1.96 & 2.84 & 0.72 & 1.67 & 1.12 & 2.58 & 1.17 \\
\rowcolor{mtoursbg}\hspace{6pt}\textbf{+ MT Forcing} & \mtwithdelta{\textbf{9.54}}{green!45!black}{+3.33} & \mtwithdelta{\textbf{8.19}}{green!45!black}{+3.78} & \mtwithdelta{\textbf{4.23}}{green!45!black}{+.68} & \mtwithdelta{\textbf{0.32}}{green!45!black}{-10.74} & \mtwithdelta{\textbf{9.55}}{green!45!black}{+5.05} & \mtwithdelta{\textbf{8.14}}{green!45!black}{+5.08} & \mtwithdelta{\textbf{4.20}}{green!45!black}{+1.06} & \mtwithdelta{\textbf{0.34}}{green!45!black}{-2.08} & \mtwithdelta{\textbf{9.16}}{green!45!black}{+6.20} & \mtwithdelta{\textbf{7.95}}{green!45!black}{+5.99} & \mtwithdelta{\textbf{4.21}}{green!45!black}{+1.37} & \mtwithdelta{\textbf{0.35}}{green!45!black}{-.37} & \mtwithdelta{\textbf{9.12}}{green!45!black}{+7.45} & \mtwithdelta{\textbf{7.94}}{green!45!black}{+6.82} & \mtwithdelta{\textbf{4.22}}{green!45!black}{+1.64} & \mtwithdelta{\textbf{0.37}}{green!45!black}{-.80} \\
\midrule
SenseNova-U1.5~\cite{diao2026sensenovau15nativeunifiedvisual} & 8.93 & 7.09 & 3.99 & 8.58 & 8.72 & 6.18 & 3.72 & 9.44 & 6.64 & 4.75 & 3.40 & 6.84 & 4.52 & 3.22 & 3.03 & 6.89 \\
\rowcolor{mtoursbg}\hspace{6pt}\textbf{+ MT Forcing} & \mtwithdelta{\textbf{9.85}}{green!45!black}{+.92} & \mtwithdelta{\textbf{8.36}}{green!45!black}{+1.27} & \mtwithdelta{\textbf{4.33}}{green!45!black}{+.34} & \mtwithdelta{\textbf{0.62}}{green!45!black}{-7.96} & \mtwithdelta{\textbf{9.66}}{green!45!black}{+.94} & \mtwithdelta{\textbf{8.32}}{green!45!black}{+2.14} & \mtwithdelta{\textbf{4.35}}{green!45!black}{+.63} & \mtwithdelta{\textbf{0.69}}{green!45!black}{-8.75} & \mtwithdelta{\textbf{9.49}}{green!45!black}{+2.85} & \mtwithdelta{\textbf{8.19}}{green!45!black}{+3.44} & \mtwithdelta{\textbf{4.34}}{green!45!black}{+.94} & \mtwithdelta{\textbf{0.76}}{green!45!black}{-6.08} & \mtwithdelta{\textbf{9.17}}{green!45!black}{+4.65} & \mtwithdelta{\textbf{7.68}}{green!45!black}{+4.46} & \mtwithdelta{\textbf{4.35}}{green!45!black}{+1.32} & \mtwithdelta{\textbf{0.71}}{green!45!black}{-6.18} \\
\bottomrule
\end{tabular*}
\end{table}

\begin{figure}[t]
    \centering
    \includegraphics[width=\linewidth]{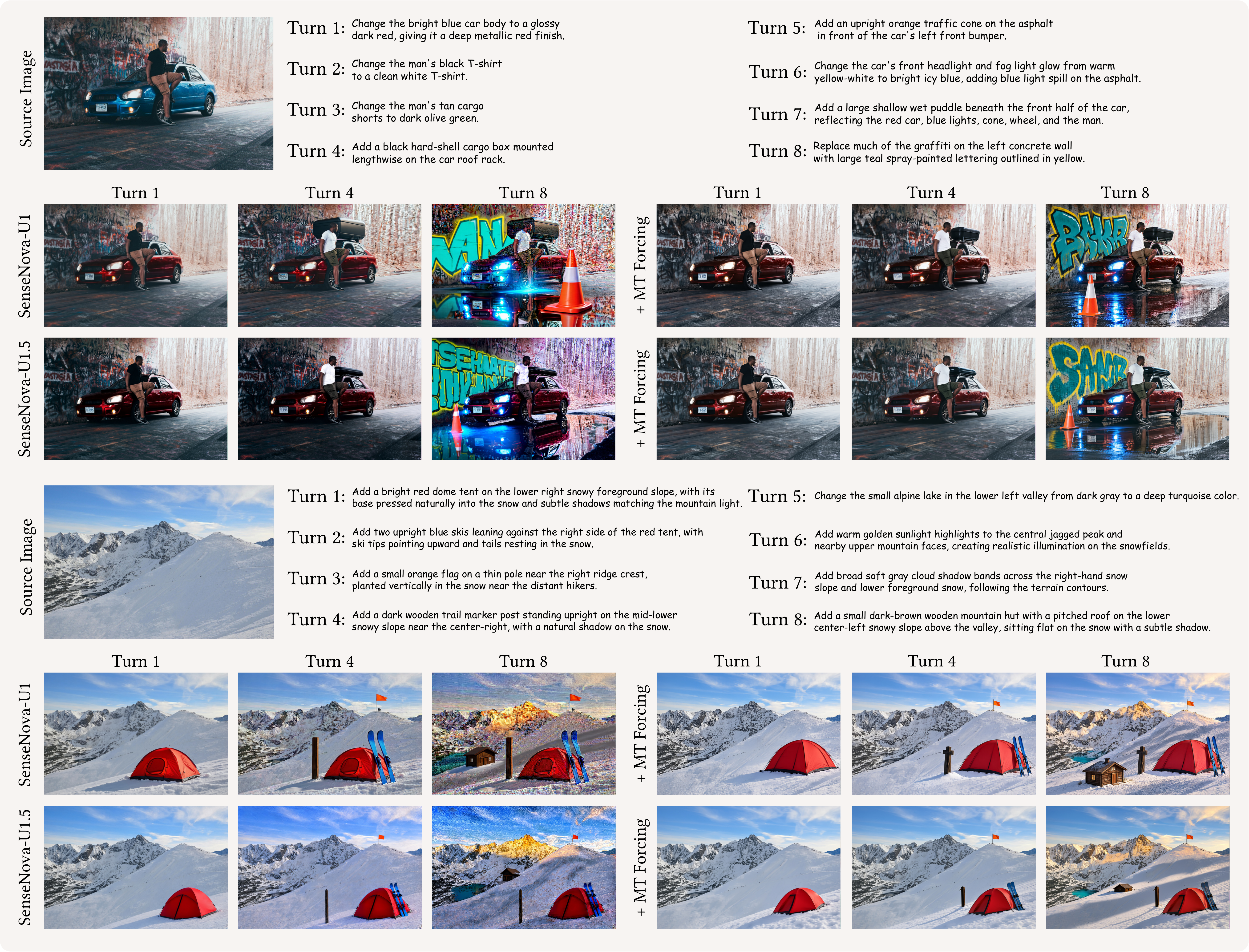}
    \caption{\textbf{Qualitative results on the step-wise sub-bench.} SenseNova-U1 and SenseNova-U1.5 are compared before (left) and after (right) MT Forcing at Turns~1, 4, and~8. MT Forcing reduces artifacts while preserving accumulated edits.}
    \label{fig:main_results}
\end{figure}

\begin{figure}[!t]
    \centering
    \includegraphics[width=\linewidth]{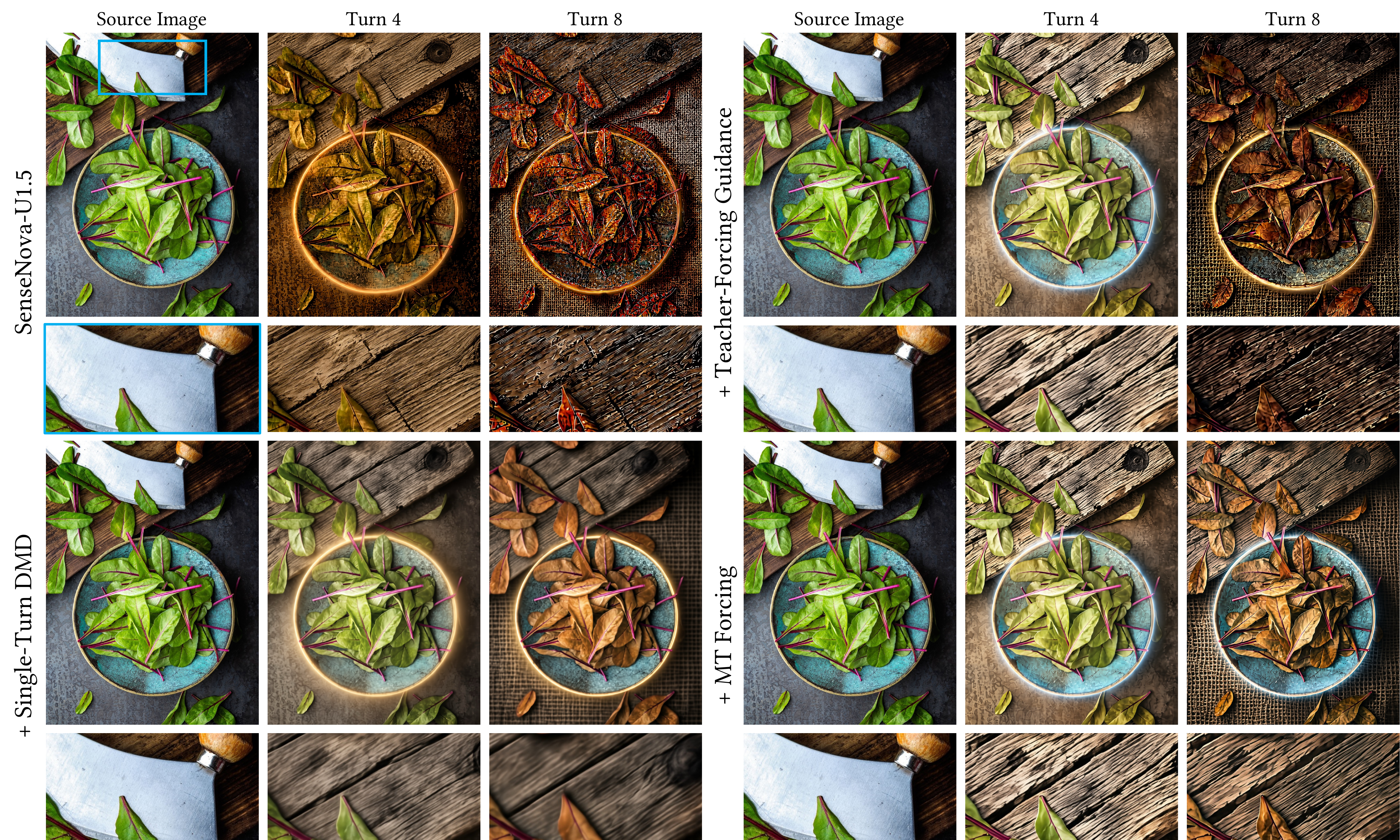}
    \caption{\textbf{Qualitative ablation of MT Forcing.} SenseNova-U1.5 is compared with Single-Turn DMD only, Teacher-Forcing Guidance only, and the full method at Turns~4 and~8. Enlarged crops highlight differences in texture detail.}
    \label{fig:ablation_components}
\end{figure}

\paragraph{Quantitative comparison.}
Table~\ref{tab:mt_step_wise_results} shows that MT Forcing outperforms both base editors on all four metrics from Turn~4 onward, with increasingly pronounced gains in IF, CC, and \(\mathrm{DRS}\) at later turns.
At Turn~8, IF and CC improve by 7.45 and 6.82 on SenseNova-U1, and by 4.65 and 4.46 on SenseNova-U1.5, respectively.
On SenseNova-U1.5, MT Forcing maintains \(\mathrm{DRS}\) between 4.20 and 4.35 throughout the session, while the baseline declines from 4.09 to 3.03.
Its eight-turn mean \(\mathrm{Pixel}\) also decreases from 6.34 to 0.74, an 88.4\% reduction in matched-point displacement.

\paragraph{Qualitative comparison.}
Fig.~\ref{fig:main_results} compares SenseNova-U1 and SenseNova-U1.5 with their MT Forcing counterparts at Turns~1, 4, and~8.
By Turn~8, the baselines exhibit oversaturated lighting and harsh textures in the car scene, and speckled snow and exaggerated contrast in the mountain scene.
MT Forcing preserves cleaner textures and coherent scene structure across turns while retaining earlier edits, including the red car body, roof box, tent, skis, and marker post.
These examples visually support the improved drift robustness and content consistency measured in Table~\ref{tab:mt_step_wise_results}.

\subsection{Ablation Studies}
\label{sec:ablation_studies}

\begin{wraptable}{r}{0.5\textwidth}
    \vspace{-6pt}
    \centering
    \captionsetup{font=footnotesize,skip=3pt}
    \caption{\textbf{Component ablation of MT Forcing.}
    Eight-turn mean scores on SenseNova-U1.5 for the step-wise sub-bench of MTEdit-Bench. The full method performs best across all four metrics.}
    \label{tab:mt_forcing_ablation}
    \begingroup
    \fontsize{7.5}{8}\selectfont
    \setlength{\tabcolsep}{2pt}
    \renewcommand{\arraystretch}{1.05}
    \begin{tabularx}{\linewidth}{cc*{4}{>{\centering\arraybackslash}X}}
        \toprule
        \shortstack{Single-Turn\\DMD}
        & \shortstack{Teacher-Forcing\\Guidance}
        & IF$\uparrow$ & CC$\uparrow$ & \(\mathrm{DRS}\uparrow\)
        & \(\mathrm{Pixel}\downarrow\) \\
        \midrule
        $\times$ & $\times$ & 8.47 & 7.09 & 3.83 & 6.34 \\
        $\checkmark$ & $\times$ & 9.02 & 7.64 & 4.12 & 2.51 \\
        $\times$ & $\checkmark$ & 9.10 & 7.83 & 3.92 & 2.74 \\
        \midrule
        \rowcolor{mtoursbg}
        $\checkmark$ & $\checkmark$ & 9.66 & 8.40 & 4.30 & 0.74 \\
        \bottomrule
    \end{tabularx}
    \endgroup
    \vspace{-6pt}
\end{wraptable}

\begin{figure}[!t]
    \centering
    \includegraphics[width=\linewidth]{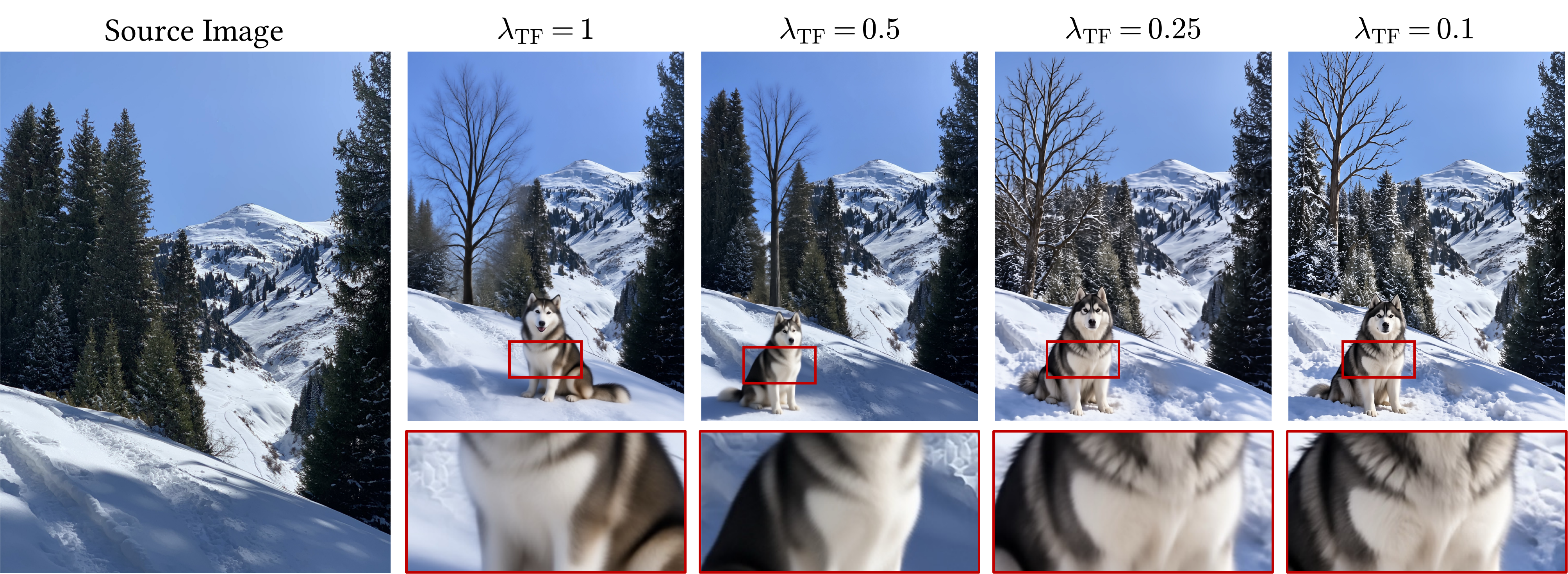}
    \caption{\textbf{Visual quality comparison under different regularization weights.} From left to right: the source image and outputs with $\lambda_{\mathrm{TF}}=1, 0.5, 0.25, 0.1$. Enlarged crops show differences in fur detail.}
    \label{fig:ablation_regularization}
\end{figure}

Table~\ref{tab:mt_forcing_ablation} and Fig.~\ref{fig:ablation_components} compare the pretrained baseline, Single-Turn DMD only, Teacher-Forcing Guidance only, and full MT Forcing.
The DMD-only variant initializes all three models from the pretrained editor and omits reference supervision.
The guidance-only variant uses teacher-forcing warmup followed by continued reference supervision, without DMD.
The full method combines warmup with DMD and teacher-forcing regularization.
We further investigate the teacher-forcing regularization weight $\lambda_{\mathrm{TF}}$ (Fig.~\ref{fig:ablation_regularization}).

\paragraph{Ablation of the two components.}
Table~\ref{tab:mt_forcing_ablation} shows that either Single-Turn DMD or Teacher-Forcing Guidance alone improves all four metrics over the pretrained baseline, while combining them yields the best results: 9.66 IF, 8.40 CC, 4.30 \(\mathrm{DRS}\), and 0.74 \(\mathrm{Pixel}\).
The qualitative comparison in Fig.~\ref{fig:ablation_components} reveals their complementary roles.
DMD alone improves visual quality by reducing the baseline's exaggerated textures, but cross-turn content inconsistency remains evident: the enlarged wood-grain patterns change between Turns~4 and~8.
Teacher-Forcing Guidance alone benefits from supervised fine-tuning on consistent reference trajectories, but training on reference histories does not address exposure bias under self-generated histories.
Consequently, visual-quality degradation persists, with harsh wood-grain textures and excessive contrast at Turn~8.
Full MT Forcing combines distribution matching on student rollouts with reference supervision, preserving more consistent wood-grain structure and clearer local details across turns.
Together, the quantitative and qualitative results support the complementary benefits of the two components for visual-quality stability and cross-turn consistency.

\paragraph{Effect of the regularization weight.}
The teacher-forcing regularization weight $\lambda_{\mathrm{TF}}$ controls the balance between reference supervision for consistency and rollout distribution matching for visual quality.
Fig.~\ref{fig:ablation_regularization} compares weights of $1$, $0.5$, $0.25$, and $0.1$.
Larger weights introduce visible blur: at $1$ and $0.5$, the husky's fur appears smooth, and individual strands are difficult to distinguish in the enlarged crops.
Reducing the weight to $0.25$ or $0.1$ restores finer fur detail and sharper coat boundaries, offering a favorable balance between content consistency and visual clarity.

\section{Conclusion}
\label{sec:conclusion}

We presented MT Forcing, which combines learning on self-generated histories with reference supervision for stable and consistent multi-turn image editing with autoregressive UMMs.
Single-Turn DMD transfers single-turn editing quality to student rollouts using a teacher that realizes cumulative editing intent directly from the source image, avoiding recursive error accumulation in its supervision.
Teacher-Forcing Guidance complements marginal distribution matching with reference-conditioned supervision during warmup and distillation, encouraging the history--output dependencies needed for content preservation.
Experiments on SenseNova-U1 and SenseNova-U1.5 demonstrate improved instruction following, content consistency, and drift robustness on the eight-turn step-wise sub-bench of MTEdit-Bench, with increasingly pronounced benefits at later turns.

\clearpage
\bibliographystyle{ieeenat_fullname}
\bibliography{main}
\end{document}